\documentclass[preprint,12pt]{elsarticle}

\usepackage{amssymb}
\usepackage{amsmath}
\usepackage{booktabs}
\usepackage{algorithm} 
\usepackage{algpseudocode}
\usepackage{hyperref} 
\usepackage{multirow} 
\usepackage{lscape}
\usepackage{hyperref}
\usepackage{xurl}
\journal{Information Sciences}

\begin{document}

\begin{frontmatter}



\title{On the Effectiveness of LLM-Specific Fine-Tuning for Detecting AI-Generated Text} 

\author[label1,label2]{Michał Gromadzki}
\author[label1]{Anna Wróblewska}
\author[label3]{Agnieszka Kaliska}

\affiliation[label1]{organization={Faculty of Mathematics and Information Science, Warsaw University of Technology},
            addressline={Plac Politechniki 1},
            city={Warsaw},
            postcode={00-661},
            country={Poland}}

\affiliation[label2]{organization={Samsung R\&D Institute Poland},
            addressline={Plac Europejski 1},
            city={Warsaw},
            postcode={00-844},
            country={Poland}}

\affiliation[label3]{organization={Faculty of Modern Languages and Literatures, Adam Mickiewicz University},
            addressline={Al. Niepodległości 4},
            city={Poznan},
            postcode={61-874},
            country={Poland}}

%% Abstract
\begin{abstract}
The rapid progress of large language models has enabled the generation of text that closely resembles human writing, creating challenges for authenticity verification in education, publishing, and digital security. Detecting AI-generated text has therefore become a crucial technical and ethical issue. This paper presents a comprehensive study of AI-generated text detection based on large-scale corpora and novel training strategies. We introduce a 1-billion-token corpus of human-authored texts spanning multiple genres and a 1.9-billion-token corpus of AI-generated texts produced by prompting a variety of LLMs across diverse domains. Using these resources, we develop and evaluate numerous detection models and propose two novel training paradigms: Per LLM and Per LLM family fine-tuning. Across a 100-million-token benchmark covering 21 large language models, our best fine-tuned detector achieves up to $99.6\%$ token-level accuracy, substantially outperforming existing open-source baselines.
\end{abstract}

\begin{graphicalabstract}
\includegraphics[width=\textwidth]{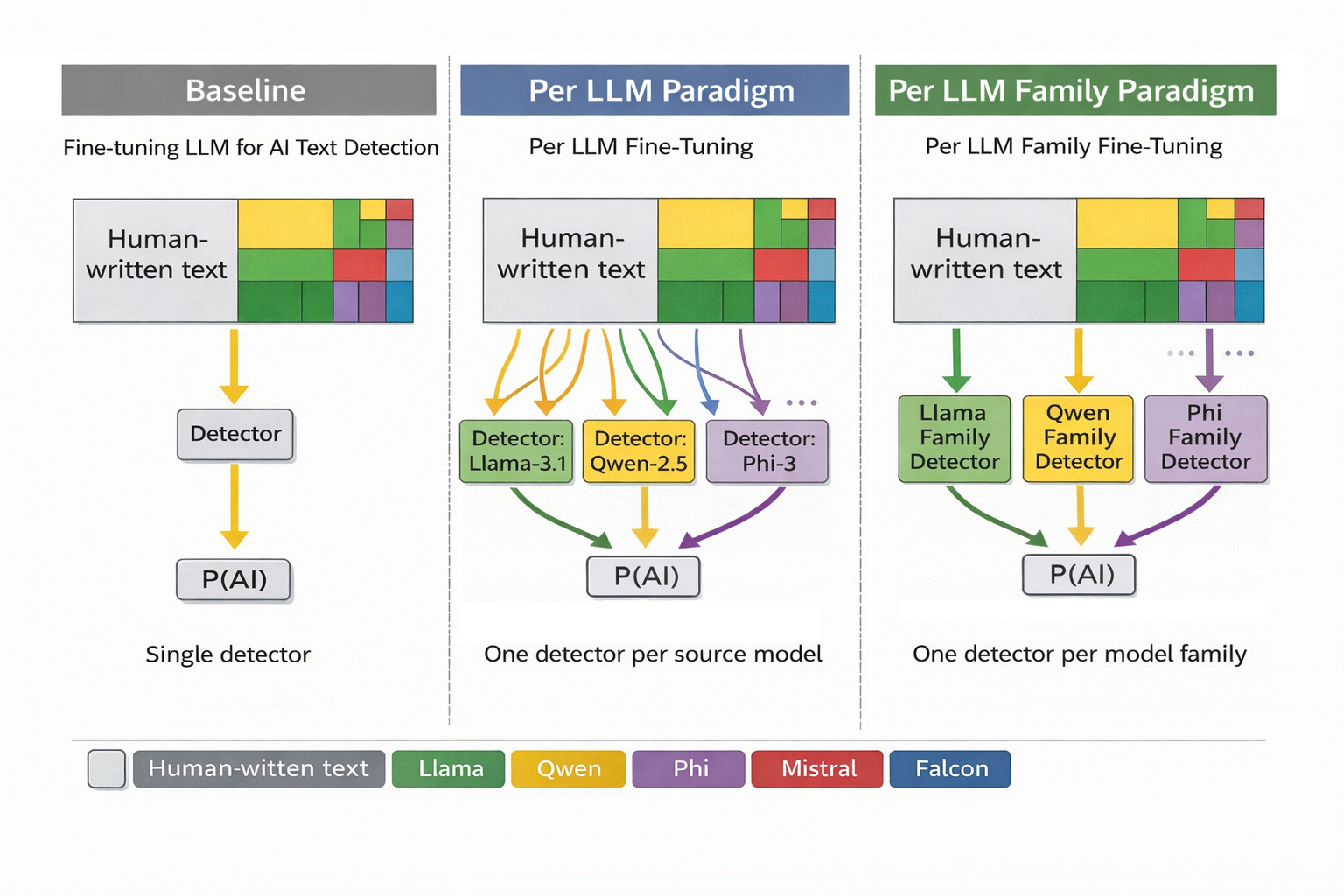}
\end{graphicalabstract}

\begin{highlights}
\item Gathered a 1-billion-token corpus of human-written texts covering multiple genres
\item Proposed a framework for generating AI-generated text across multiple language models
\item Generated a 1.9-billion-token corpus of AI-generated texts from 21 language models
\item Introduced two novel training paradigms for AI-generated text detection
\item Conducted extensive experiments in AI-generated text detection
\end{highlights}

%% Keywords

\end{frontmatter}



\section{Introduction} \label{sec:introduction}

Recent advances in natural language processing have produced large language models capable of generating text that closely resembles human writing. While models such as OpenAI’s GPT series, Google’s Gemini, and Anthropic’s Claude offer unprecedented opportunities in domains like customer support, journalism, and creative writing, they also pose challenges in verifying authenticity and assessing text quality. Distinguishing between AI-generated and human-authored text has become a critical technical and ethical issue, particularly in education, publishing, and digital security.

This work makes several significant contributions to the field of AI-generated text detection. First, we gathered a 1-billion-token corpus of human-written texts spanning a wide range of genres. We then proposed a scalable framework for generating AI-generated text at scale across multiple large language models. Using this framework, we generated a corpus of 1.9 billion tokens of AI-generated text from 21 large language models across diverse domains. We further proposed a precise sampling algorithm to construct balanced experimental datasets from these corpora. Finally, we conducted extensive experiments with multiple LLMs, introducing two novel training paradigms for AI-generated text detection: fine-tuning \textbf{per LLM} and \textbf{per LLM family}. The code is available at \href{https://github.com/Michal1337/llm-ai-text-detection}{github.com/Michal1337/llm-ai-text-detection}

\section{Related Work} \label{sec:related_works}

\subsection{Threats Imposed By AI}

Although AI-generated texts are often considered well-structured and grammatically correct, they are also frequently viewed as neutral or impersonal because they rely on statistically likely phrasing, making them more predictable than human-written texts. However, the rapid progress of large language models such as GPT~\cite{openai2024gpt4technicalreport}, Gemini~\cite{geminiteam2024geminifamilyhighlycapable}, and Claude~\cite{AnthropicClaude2023} has made it increasingly challenging to distinguish human-written text from machine-generated content~\cite{tang2023sciencedetectingllmgeneratedtexts, wu2025surveyLLMdetection}. While these systems have unlocked powerful applications in natural language understanding and generation, their widespread use raises concerns regarding academic integrity, misinformation, spam, and malicious disinformation campaigns~\cite{weber_wulff2023testing}. 

AI-generated text poses multiple threats across societal, academic, political, and ethical aspects. One of the most concerning issues is the spread of misinformation and disinformation. Because modern LLMs produce fluent, coherent text that can mimic human writing very convincingly, they can be used to generate fake news, propaganda, or misleading health advice~\cite{Zao-Sanders2025GenAI}. Once this kind of content spreads, especially on social media, it can shape opinions, foster mistrust, and even put people's health and safety at risk. The ability to produce large volumes of reasonable text quickly amplifies this: adversaries can flood information channels with propaganda or false claims, making it difficult for fact-checkers and the public to distinguish truth from fabrication~\cite{shoaib2023deepfakes}.

In academic settings, AI-generated text challenges academic integrity and assessment. Students may use LLMs to produce essays, reports, or homework assignments that appear authentic but are not their own work. Studies show that existing detection tools are imperfect. They often have high rates of false negatives (fail to detect AI-written text) or false positives (flag human-written text as AI). It undermines confidence in detection tools and academic policies based on them~\cite{sadasivan2023can, weber_wulff2023testing}.

Another threat arises from adversarial robustness and evasion. As detection tools become more available, adversaries can use paraphrasing, style changes, translation, or other transformations to obscure AI generation and bypass detectors~\cite{sadasivan2023can}. For instance, recursive paraphrasing has been shown to significantly reduce detection rates while preserving readability. This arms-race dynamic means that any detection method must continuously adapt~\cite{sadasivan2023can}.

Finally, there are legal and regulatory risks. Accusations of using AI may have serious consequences, especially if based on flawed detection. If institutions rely on opaque or unreliable tools, this may lead to unfair sanctions, reputational harm, and possible legal challenges~\cite{weber_wulff2023testing, guardian2024aicrisis}. Furthermore, the fast pace of generative AI development often outpaces oversight and regulation, leaving gaps in accountability~\cite{weber_wulff2023testing,rosca2025new}.

\subsection{Detection Methods}

Statistical and stylometric methods are among the earliest and still most widely used approaches to detecting machine-generated text. These techniques rely on token-level statistics, such as perplexity or log-probabilities under a language model, as well as n-gram rank distributions, burstiness, and other stylometric features. Tools like GLTR\footnote{gltr.io} make these signals accessible to human analysts through visualization and inspection~\cite{gehrmann2019gltr}. These methods are attractive because they are simple, interpretable, and computationally inexpensive, and they often perform well against older generators or text produced with naive decoding strategies. However, modern LLMs generate low-perplexity, highly fluent text, and decoding choices (e.g., temperature, top-k) alter statistical footprints in ways that can make detection more difficult~\cite{ippolito2020automatic,gehrmann2019gltr}.

Supervised classifiers and fine-tuned detectors are a common, high-performing approach for distinguishing human-written from machine-generated text. These methods typically involve training a discriminator, often a fine-tuned transformer such as RoBERTa or BERT, on labeled examples of human and machine-generated text. Variants include training on diverse synthetic sources (multi-model training), which forms the backbone of many commercial detectors and research baselines. Earlier work even showed that training a generator and then using that generator itself as a detector can be surprisingly effective~\cite{zellers2019grover}. Fine-tuned classifiers perform exceptionally well when the training distribution closely matches the evaluation distribution. However, they frequently overfit to dataset- or model-specific artifacts and degrade when faced with unseen generators, new domains, or paraphrasing attacks~\cite{wu2025surveyLLMdetection,sadasivan2023can}.

Model-aware and zero-shot detectors, such as DetectGPT and its derivatives, provide an alternative to training separate classifiers by leveraging the properties of the generator itself. A canonical example is DetectGPT, which leverages the local curvature properties of a generator's log-probability function. Machine-sampled text tends to occupy regions of negative curvature under the generating model's log-probability, and detection is performed by sampling perturbations and measuring this curvature. This approach can operate without a labeled dataset but requires access to, or the ability to query, the generating model~\cite{mitchell2023detectgpt}. Fast variants, such as Fast-DetectGPT, reduce the computational burden. Yet, these methods remain more expensive than simple perplexity-based checks and may be impractical when the generator is inaccessible or multiple models must be covered~\cite{mitchell2023detectgpt,bao2023fastdetectgpt}.

Watermarking and provenance techniques offer a proactive approach to detecting machine-generated text by embedding detectable, yet invisible to humans, signals during the generation process. When the generator cooperates, practical watermarking schemes demonstrate strong detectability with minimal degradation of text quality~\cite{kirchenbauer2023watermark}. Subsequent work has examined the robustness of these watermarks against paraphrasing, insertion into longer documents, or human rewriting: detection remains possible given sufficient token coverage, but robustness varies with attack strength and whether watermarks are universally adopted~\cite{kirchenbauer2023watermark, kirchenbauer2023reliability}.

\section{Our Approach}

This study presents a systematic experimental framework that serves as a reliable benchmark for AI-generated text detection. The framework supports both the evaluation of detection methods and the construction, maintenance, and extension of datasets. Such capabilities are increasingly important as large language models continue to improve at generating fluent, diverse text, posing ongoing challenges for robust detection.

Thus, in the following, we first describe our data pipeline for generating text in Section~\ref{sec:data_pipeline}. Then, in Section~\ref{sec:experimental_setup}, we present our experimental setup to test the ability of various LLMs to differentiate between human and generated text. Next, we provide a precise description of each experiment in Section~\ref{sec:experiments}, in which we also introduce our fine-tuning paradigms: \textbf{per LLM} and per \textbf{LLM family}.
Then, we report and analyze the results in Section~\ref{sec:results}.

\section{Data Pipeline} \label{sec:data_pipeline}

\begin{figure}
    \centering
    \includegraphics[width=\textwidth]{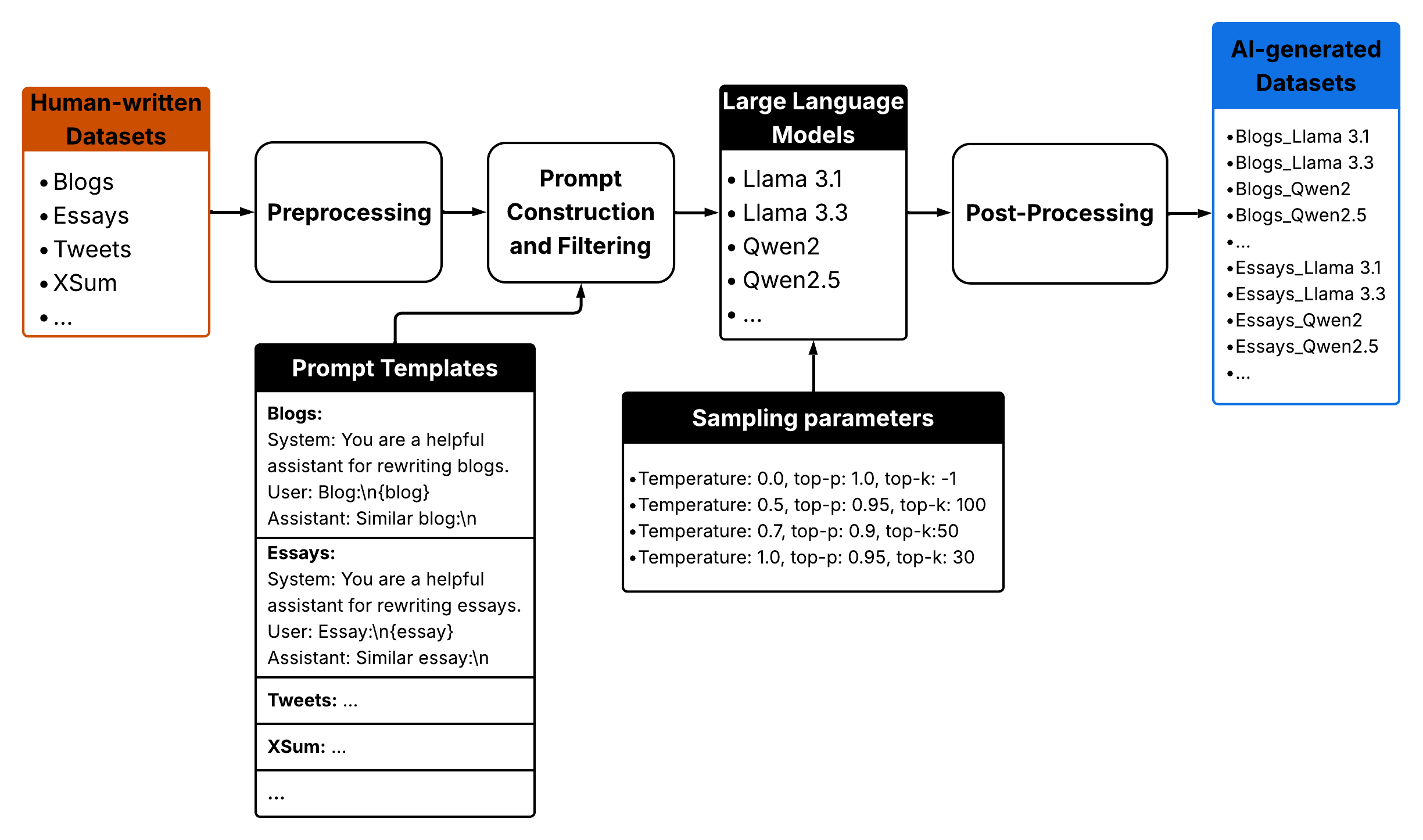}
    \caption{Overview of the data generation pipeline}
    \label{fig:data_pipeline}
\end{figure}

The data consists of human-authored and AI-generated texts. We describe the approach to generating AI-made texts and explain the selection of large language models. This data serves as the foundation for the subsequent analysis and experiments in AI text detection. Figure~\ref{fig:data_pipeline} presents the overview of our data pipeline.

\subsection{Human-authored Corpora} \label{ssec:htext}

We have selected $10$ human-written textual datasets\footnote{Data Sources:~\ref{app:ds}}. The human-authored texts encompass a range of genres, including blogs, essays, news articles, online discussions, and creative writing. Due to the versatility of large language models, preprocessing is usually limited to removing samples that are too short or too long, discarding unused columns, and deduplicating datasets.

\subsection{AI-generated Corpora} \label{ssec:aitext}

The AI-generated texts were obtained through controlled text generation using 20 popular local large language models and one proprietary model.

\subsubsection{Overview}

We generated AI-produced text by conditioning large language models on human-written samples. For each selected human dataset, a prompt was constructed for every sample using a predefined template. Prompts exceeding $16{,}384$ tokens were discarded, and, for larger datasets, a random subset of the remaining prompts was sampled. The selected prompts were then batched and passed to the LLM, with one of four predefined sampling parameter configurations randomly assigned to each batch. After generation, the outputs were post-processed to improve text quality. This procedure was applied consistently across all datasets and language models.

\subsubsection{Large Language Model Selection}

Table~\ref{tab:model_info} provides company names, parameter counts, and quantization types of selected large language models.

\begin{table}[H]
    \centering
    \small % Optional: makes table more compact
    \resizebox{\textwidth}{!}{
    \begin{tabular}{llcc}
        \toprule
        \textbf{Company} & \textbf{Model Name} & \textbf{Param \#} & \textbf{Quant} \\
        \midrule
        OpenAI & gpt-4.1-nano-2025-04-14 & N/A & N/A \\
        \midrule
        Meta & Llama-3.1-8B-Instruct & 8B & -- \\
        Meta & Meta-Llama-3.1-70B-Instruct-AWQ-INT4 & 70B & AWQ \\
        Meta & Llama-3.2-3B-Instruct & 3B & -- \\
        Meta & Meta-Llama-3.3-70B-Instruct-AWQ-INT4 & 70B & AWQ \\
        \midrule
        Microsoft & Phi-3-mini-128k-instruct & 4B & -- \\
        Microsoft & Phi-3-small-128k-instruct & 7B & -- \\
        Microsoft & Phi-3-medium-128k-instruct & 14B & -- \\
        Microsoft & Phi-3.5-mini-instruct & 4B & -- \\
        Microsoft & Phi-4-mini-instruct & 4B & -- \\
        Microsoft & Phi-4 & 14B & -- \\
        \midrule
        Mistral AI & Ministral-8B-Instruct-2410 & 8B & -- \\
        Mistral AI & Mistral-Nemo-Instruct-2407 & 12B & -- \\
        \midrule
        Alibaba Cloud & Qwen2-7B-Instruct & 7B & -- \\
        Alibaba Cloud & Qwen2-72B-Instruct-AWQ & 72B & AWQ \\
        Alibaba Cloud & Qwen2.5-3B-Instruct & 3B & -- \\
        Alibaba Cloud & Qwen2.5-7B-Instruct & 7B & -- \\
        Alibaba Cloud & Qwen2.5-14B-Instruct & 14B & -- \\
        Alibaba Cloud & Qwen2.5-72B-Instruct-AWQ & 72B & AWQ \\
        \midrule
        TII & Falcon3-3B-Instruct & 3B & -- \\
        TII & Falcon3-7B-Instruct & 7B & -- \\
        \bottomrule
    \end{tabular}}
    \caption{Large language model information}
    \label{tab:model_info}
\end{table}

The models were selected to enable multiple comparisons in the latter parts of our work. All models support a context length of at least $16\;384$ tokens.

\subsubsection{Prompts}

To create a prompt, specified features from a sample in the selected human dataset were inserted into the prompt template. This approach resulted in a one-to-one mapping between human dataset samples and the corresponding prompts and, later, between human-written and AI-generated samples. Prompt templates for all datasets are provided in~\ref{app:pt}.

All prompt templates consist of three messages. The first message outlines the task of the generative model. In our research, the tasks are limited to rewriting provided texts or writing new texts based on provided features. The second message provides the model with the necessary information to complete the task specified in the system message, such as rewriting a blog post or summarizing an article. The final message is the beginning of the model's response. Presetting the start of the model's response trivialized the process of extracting the actual text of a completed task, such as a rewritten blog, from the entire model's response. This prompt template structure ensured that all instruction fine-tuned generative models could have been used to generate AI text samples.

\subsubsection{Sampling parameters}

Along with prepared prompts, we passed sampling parameters to the generative models. We randomly selected one of four preset sampling parameter combinations for each prompt batch. Doing so increased the diversity of the generated text. Table~\ref{tab:sampling_params} provides the sampling parameters presets.

\begin{table}[H]
    \centering
    \begin{tabular}{cccc}
        \toprule
        \textbf{Preset} & \textbf{Temperature} & \textbf{Top-p} & \textbf{Top-k} \\
        \midrule
        Deterministic & 0.0 & 1.0 & -1 \\
        Balanced & 0.5 & 0.95 & 100 \\
        Creative & 0.7 & 0.9 & 50 \\
        Highly Creative & 1.0 & 0.95 & 30 \\
        \bottomrule
    \end{tabular}
    \caption{Sampling parameter presets}
    \label{tab:sampling_params}
\end{table}

The three key sampling parameters are:
\begin{itemize}
    \item \textbf{Temperature}: Controls output randomness; higher values increase diversity, while lower values yield more deterministic generations.
    \item \textbf{Top-p}: Restricts sampling to the smallest set of tokens whose cumulative probability exceeds $p$, balancing diversity and coherence.
    \item \textbf{Top-k}: Limits sampling to the $k$ most probable next tokens, reducing unlikely outputs while preserving diversity.
\end{itemize}

\subsubsection{Post-processing}

Post-processing was necessary to address issues in a small subset of AI-generated texts. Some samples contained extremely long sequences with excessive whitespace (e.g., repeated spaces or newlines) or repeated phrases. These unusual outputs were likely generated by weaker models operating at high temperature settings, further degrading their output quality. To address those issues, we applied a post-processing algorithm that cleans and normalizes each AI-generated sample. 

\subsection{Summary of Gathered Corpora}

We have acquired a total of $220 = 10 \text{ (datasets)}\;\cdot\;22 \text{ (21 LLMs + human)}$ different textual datasets.

Table~\ref{tab:summary_agg_data} provides the corpora statistics aggregated by the human dataset.

\begin{table}[H]
\centering
\begin{tabular}{@{}lrrrr@{}}
\toprule
\textbf{Human Dataset} & \textbf{\# Samples} & \textbf{\# Sentences} & \textbf{\# Words} & \textbf{\# Tokens} \\
\midrule
Blogs            & $1.2$M  & $21.2$M  & $381$M  & $407$M \\
Essays           & $58$K   & $2.8$M   & $40.2$M & $39.4$M \\
Natural Questions& $474$K  & $2.3$M   & $46.4$M & $58.8$M \\
NYT Articles     & $348$K  & $2.2$M   & $60.1$M & $64.0$M \\
NYT Comments     & $8.7$M  & $36.3$M  & $706$M  & $747$M \\
RAID             & $864$K  & $10.8$M  & $264$M  & $333$M \\
Reddit           & $3.4$M  & $14.0$M  & $296$M  & $340$M \\
Tweets           & $3.7$M  & $8.1$M   & $97.2$M & $118$M \\
WritingPrompts   & $621$K  & $23.3$M  & $379$M  & $399$M \\
XSum             & $940$K  & $12.8$M  & $340$M  & $360$M \\
\midrule
\textbf{Total}   & $\mathbf{20.2}$\textbf{M} & $\mathbf{133.8}$\textbf{M} & $\mathbf{2.6}$\textbf{B} & $\mathbf{2.9}$\textbf{B} \\
\bottomrule
\end{tabular}
\caption{Corpora statistics aggregated by the human datasets}
\label{tab:summary_agg_data}
\end{table}

Table~\ref{tab:summary_agg_is_human} provides the corpora statistics aggregated by the source of the texts.

\begin{table}[H]
\centering
\begin{tabular}{@{}lrrrr@{}}
\toprule
\textbf{Source} & \textbf{\# Samples} & \textbf{\# Sentences} & \textbf{\# Words} & \textbf{\# Tokens} \\
\midrule
Human-written & $6.8$M & $50.2$M & $886$M & $1.0$B \\
AI-generated  & $13.5$M & $83.6$M & $1.7$B & $1.9$B \\
\midrule
\textbf{Total} & $\mathbf{20.2}$\textbf{M} & $\mathbf{133.8}$\textbf{M} & $\mathbf{2.6}$\textbf{B} & $\mathbf{2.9}$\textbf{B} \\
\bottomrule
\end{tabular}
\caption{Corpora statistics aggregated by the source}
\label{tab:summary_agg_is_human}
\end{table}

Overall, the gathered corpora cover a broad range of domains, genres, and text lengths. AI-generated text constitutes a larger share of the data due to the inclusion of multiple language models per dataset. At the same time, the diversity of datasets ensures coverage of both short-form and long-form writing. This scale and diversity support a comprehensive evaluation of AI-generated text detection methods and their ability to generalize across domains and model families. The entire corpus is available at \href{https://huggingface.co/datasets/Majkel1337/Detect-AI}{huggingface.co/datasets/Majkel1337/Detect-AI}

\subsection{Experimental Datasets Construction}

Experimental datasets are labeled subsets of our entire corpus. All datasets were prepared for use in binary classification tasks. Each sample in the dataset is a pair of text, written by a human or generated by AI, and a label indicating whether the text was written by a human or generated by AI.

\begin{figure}[H]
    \centering
    \includegraphics[width=\textwidth]{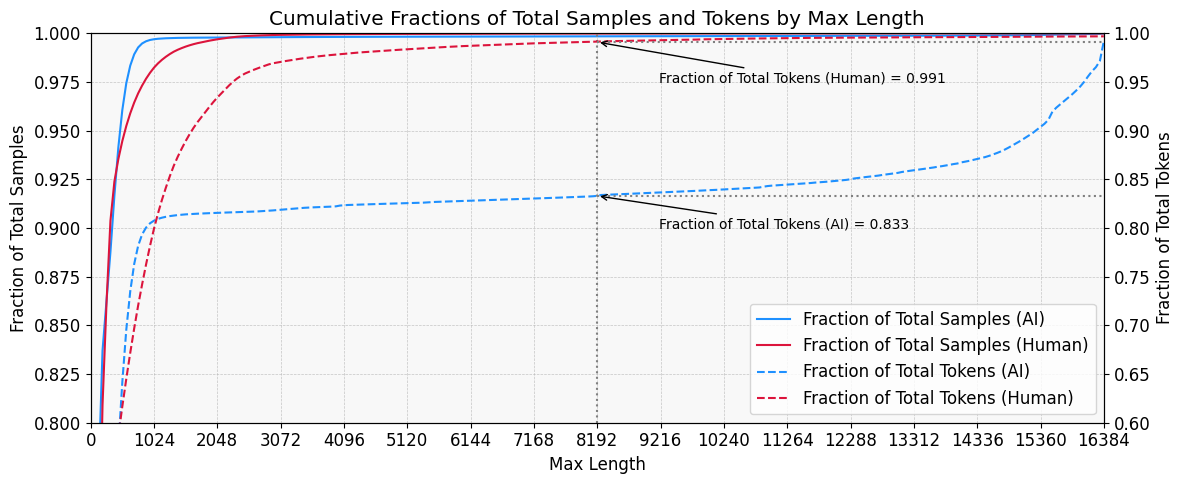}
    \caption{Cumulative fractions of total samples and tokens by max length}
    \label{fig:cutoff}
\end{figure}

During the construction of the datasets, we considered only texts of up to $8192$ tokens in length. Figure~\ref{fig:cutoff} presents the cumulative fractions of total samples and tokens by max length for human-written and AI-generated texts.

The cutoff of $8192$ tokens removed only a negligible number of samples from both human-written and AI-generated datasets. While the fraction of total removed tokens is significantly higher, we still keep over $99\%$ of human-written tokens and over $83\%$ of AI-generated tokens.

During the construction of all experimental datasets, we met the following requirements:
\begin{itemize}
    \item Datasets must be balanced between both classes in token counts
    \item Datasets must be balanced between generative models in token counts, in both classes
    \item Number of tokens of each text genre must be proportional to the total collected tokens in the selected text genre
    \item Number of tokens in the validation split is equal to approximately $30\%$ of training tokens
\end{itemize}

These requirements ensure that each experimental dataset is representative and suitable for rigorous evaluation of AI-generated text detection methods. By balancing token counts across classes, models, and genres and maintaining a consistent validation split, we aim to minimize potential biases and enable fair comparisons between detection approaches.

Table~\ref{tab:datasets} presents the total number of samples, sentences, and tokens in each dataset.

\begin{table}[H]
\centering
\begin{tabular}{@{}llrrr@{}}
\toprule
\textbf{Dataset} & \textbf{Split} & \textbf{Samples} & \textbf{Sentences} & \textbf{Tokens} \\
\midrule
\multirow{2}{*}{\textit{master-large}} 
  & Train & $366$K & $5$M & $100$M \\
  & Validation & $115$K & $1.5$M & $30$M \\
\midrule
\multirow{1}{*}{\textit{master-testset}} 
  & Test & $358$K & $5$M & $100$M \\
\bottomrule
\end{tabular}
\caption{\textit{Master} datasets statistics}
\label{tab:datasets}
\end{table}

\section{Experimental Setup} \label{sec:experimental_setup}

Each model was trained to classify individual tokens in a text as either human-written or AI-generated. The idea behind token-level classification was that, in real-world scenarios, texts often contain a mix of human-written and AI-generated content. By performing token-level classification, we hoped that the models would accurately identify AI-generated text and make more meaningful overall predictions. One key requirement of our approach was that each prediction be made based on the entire text sample, regardless of which token was being classified. This requirement, along with token-level classification, was assumed in all experiments.

\subsection{Model Architecture}

\begin{figure}[H]
    \centering
    \includegraphics[width=\textwidth]{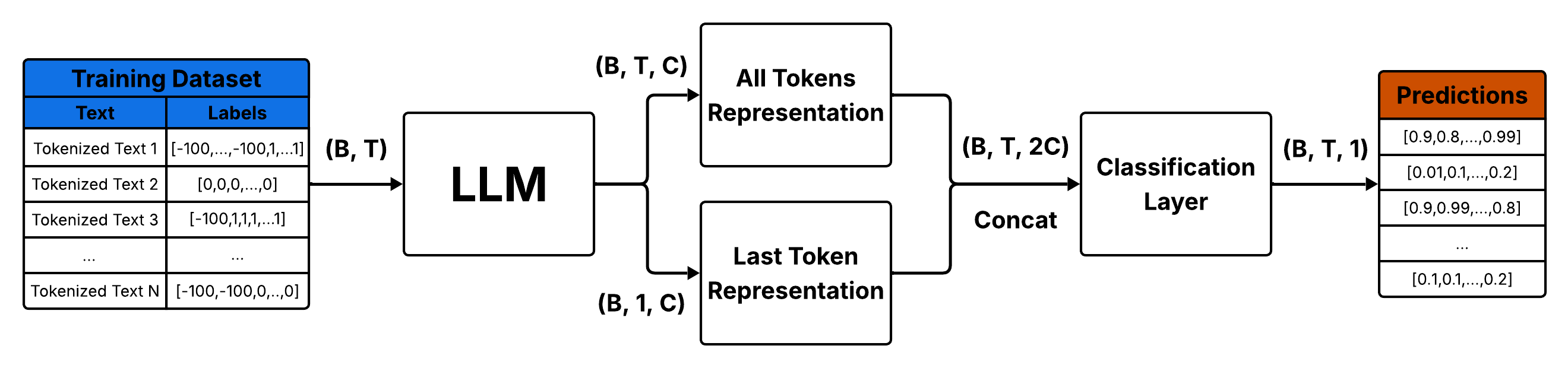}
    \caption{AI-generated text detection pipeline for LLM fine-tuning}
    \label{fig:pretrain_model}
\end{figure}

Figure~\ref{fig:pretrain_model} presents the AI-generated text detection pipeline for all of our experiments. Above the arrows, we present shapes of the resulting tensors. The batch size is denoted as $B$, text length as $T$, and $\text{d}_\text{model}$ parameter as $C$.

The LLM processes a batch of tokenized texts and produces contextualized token representations of shape $(B, T, C)$. As the model is autoregressive, each token representation encodes only left-context information. To incorporate global context into token-level predictions, we extract the representation of the final token, which summarizes the entire input sequence, and concatenate it with each token representation along the feature dimension. This operation ensures that each token-level prediction incorporates information from the full text sample. The resulting tensor of shape $(B, T, 2C)$ is then passed to a linear classification head with a sigmoid activation function.

\subsection{Model Training}

All LLMs were fine-tuned for five epochs using the Adam optimizer ($\beta_1=0.9$, $\beta_2=0.999$) and a binary cross-entropy loss function. During fine-tuning, the language model backbone was frozen, and only the final classification layer was updated. For each training run, we saved the weights of the classification layer that achieved the highest accuracy on the validation split of the corresponding dataset. A linear learning rate schedule with a $20\%$ warm-up was used. The exact hyperparameters are provided in Appendix~\ref{app:tsd}.

Before fine-tuning, all texts were padded from the left. This padding strategy ensures that the representation of the final token corresponds to an actual text token rather than padding. Token-level predictions corresponding to padding tokens were masked and excluded from loss and accuracy computations.

\section{Conducted Experiments} \label{sec:experiments}

We evaluate various AI-generated text detection approaches and investigate fine-tuning of large language models, introducing two novel training paradigms: \textbf{Per LLM} and \textbf{Per LLM family} fine-tuning. For each paradigm, we describe the construction of training datasets and analyze their strengths and limitations. The experiments further extend to detecting content generated by a proprietary LLM. Finally, we present the evaluation framework used to compare model performance and assess detection effectiveness across experiments.

\subsection{LLM Fine-tuning for AI Text Detection} \label{ssec:exp1}

We fine-tuned 16 selected local large language models (Table~\ref{tab:model_info}, excluding the four $70$B$+$ LLMs) on the \textit{master-large} training split. This experiment serves as a baseline for the two new paradigms introduced in the following experiments.

\subsection{LLM Fine-tuning for Proprietary Model Detection} \label{ssec:exp2}

In the previous experiment, we focused on detecting AI-generated texts from multiple generative models. However, in real-world scenarios, the majority of AI-generated content is produced by a small number of proprietary models and is typically accessed via web applications such as \href{https://chatgpt.com/}{chatgpt.com}~\cite{openai2024gpt4technicalreport}. In this experiment, we focused on detecting the AI-generated texts only from the \texttt{gpt-4.1-nano-2025-04-14} proprietary model.

We fine-tuned 16 selected local large language models~(Table \ref{tab:model_info}, excluding the four $70$B$+$ LLMs) on the \textit{detect-gpt-4.1-nano} training split.

\subsubsection{Training and Validation Dataset Construction}

For this experiment, we constructed a separate dataset - \textit{detect-gpt-4.1-nano}. This dataset was created with the same algorithm as the \textit{master} dataset. It resulted in a dataset in which $50\%$ of the tokens are from texts generated by the \textit{detect-gpt-4.1-nano} model and the other $50\%$ from different sources, including texts generated by other LLMs and human-written content. Table~\ref{tab:ds_detect_gpt} presents the total number of samples, sentences, and tokens in the created dataset.

\begin{table}[H]
\centering
\begin{tabular}{@{}llrrr@{}}
\toprule
\textbf{Dataset} & \textbf{Split} & \textbf{Samples} & \textbf{Sentences} & \textbf{Tokens} \\
\midrule
\multirow{2}{*}{\textit{detect-gpt-4.1-nano}} 
  & Train & $308$K & $2.4$M & $50$M \\
  & Validation & $157$K & $726$K & $15$M \\
\bottomrule
\end{tabular}
\caption{\textit{Detect-gpt-4.1-nano} dataset statistics}
\label{tab:ds_detect_gpt}
\end{table}

\subsection{Per LLM Fine-Tuning for AI Text Detection} \label{ssec:exp3}

In this experiment, we fine-tuned each of the 20 selected local large language models~(Table \ref{tab:model_info}) to detect texts generated by themselves. We hypothesize that detecting texts generated by a single, known LLM is easier than identifying AI-generated texts when they could have originated from any of several LLMs. Especially, if precisely the same large language model generated the texts and is fine-tuned to detect them. This selected LLM's weights already steer it towards generating texts of a specific style; hence, it should be the most effective among any LLMs for detecting itself. We named this approach \textbf{Per LLM} fine-tuning for AI text detection. 

In the context of AI-generated text detection, we consider a sample to be human-written if all fine-tuned LLMs classify most of its tokens as not generated by themselves. The 20 fine-tuned LLMs cover all but 1 of the generative models we have used. The missing LLM is \texttt{gpt-4.1-nano-2025-04-14}. As we do not have access to its weights, we cannot fine-tune it to detect itself. This issue was resolved by adding the best-performing detector from the previous experiment to the ensemble. In summary, the ensemble consists of 20 fine-tuned LLMs, which detect texts generated by themselves and the best-performing LLM (from the previous experiment), fine-tuned to detect texts from the proprietary \texttt{gpt-4.1-nano-2025-04-14} LLM. Those 21 fine-tuned LLMs cover all generative models used in our research.

This paradigm offers several advantages: it enables targeted optimization for a specific model and has the potential to significantly improve accuracy compared to standard detection pipelines. However, it also comes with limitations. It does not generalize to new, unseen LLMs, and, most notably, it is computationally unsustainable. While for our selected 20 local large language models, it is still feasible to fine-tune all of them. As of May 2025, there are over $700\;000$ LLMs available on \href{https://huggingface.co/models?pipeline_tag=text-generation&sort=trending}{HuggingFace}. Fine-tuning each of the $700\;000$ available LLMs would require an impossible amount of compute. Second, passing a single text sample into over $700\;000$ different fine-tuned AI-generated text detectors seems unreasonable. Those limitations were addressed in the following experiment.

We fine-tuned all 20 selected local large language models~(Table \ref{tab:model_info}) on the matching detection datasets. This experiment resulted in an ensemble of 20 fine-tuned LLMs, which serve as detectors of texts generated by themselves.

\subsubsection{Training and Validation Dataset Construction}

We created separate datasets for each of the 20 local large language models. The datasets were constructed similarly to the \textit{detect-gpt-4.1-nano} dataset. In each dataset, $50\%$ of the tokens come from texts generated by the detected LLM, while the remaining $50\%$ come from all other sources. Each of the 20 created datasets includes a training split with $60$M tokens and a validation split with $18$M tokens.

\subsection{Per LLM Family Fine-Tuning for AI Text Detection} \label{ssec:exp4}

This experiment addresses the limitations of our \textbf{Per LLM} fine-tuning paradigm. The main limitation of \textbf{Per LLM} fine-tuning paradigm is that it is extremely computationally expensive during training and inference. Instead of fine-tuning LLMs to detect texts from themselves, we propose fine-tuning them to detect texts from all models within their family. For example, each LLM from the \texttt{Llama} family was fine-tuned to detect texts from any model from the \texttt{Llama} family. 

The number of LLM families is significantly smaller than the total number of individual LLMs. For our research, which covers 20 local large language models from $5$ families, this approach reduced the number of required detectors from $20$ to $5$. It significantly reduced the compute requirement during inference. Moreover, as LLMs are fine-tuned to detect texts from all models in their families, they may generalize better to new models from those families. We named this approach \textbf{Per LLM family} fine-tuning for AI text detection.

In the context of AI-generated text detection, we first selected the best-performing detection model for each LLM family. It resulted in 5 fine-tuned LLMs that cover all 20 local large language models used. Similarly to the \textbf{Per LLM} approach, the remaining uncovered model is \texttt{gpt-4.1-nano-\allowbreak 2025-04-14}. Again, we added the best-performing text detector from the \texttt{gpt-4.1-nano-2025-04-14} model to the ensemble. In summary, the ensemble consists of $5$ fine-tuned LLMs, which detect texts generated from models in their respective families and the best-performing LLM (the same as in the previous experiment), fine-tuned to detect texts from the proprietary \texttt{gpt-4.1-nano-2025-04-14} LLM. Those 6 fine-tuned LLMs cover all generative models used in our research.

We fine-tuned 16 selected local large language models~(Table \ref{tab:model_info}, excluding the four $70$B$+$ LLMs) on the matching detection datasets training splits. This experiment produced 16 fine-tuned LLMs that serve as detectors of text generated by models from their respective families. Based on the performance of the models on their respective validation splits, we have selected one detector for each of the five LLM families that were included in the final ensemble.

\subsubsection{Training and Validation Dataset Construction}

For each LLM family (\texttt{Llama}, \texttt{Phi}, \texttt{Mistral}, \texttt{Qwen}, \texttt{Falcon}), we created separate datasets. These datasets were created with the same algorithm as the \textit{master} dataset. As a result, each dataset contains $50\%$ of tokens evenly allocated across LLMs from the target family, with the remaining $50\%$ coming from all other generative models and human-written texts. Table~\ref{tab:llm_family_ds} presents the total number of samples, sentences, and tokens in the created datasets.

\begin{table}[H]
\centering
\begin{tabular}{@{}llrrr@{}}
\toprule
\textbf{Dataset} & \textbf{Split} & \textbf{Samples} & \textbf{Sentences} & \textbf{Tokens} \\
\midrule
\multirow{2}{*}{\textit{detect-llama-family}} 
  & Train & $403$K & $4.5$M & $100$M \\
  & Val   & $127$K & $1.3$M & $30$M \\
\midrule
\multirow{2}{*}{\textit{detect-phi-family}} 
  & Train & $403$K & $4.6$M & $100$M \\
  & Val   & $128$K & $1.4$M & $30$M \\
\midrule
\multirow{2}{*}{\textit{detect-mistral-family}} 
  & Train & $527$K & $4.9$M & $100$M \\
  & Val   & $197$K & $1.4$M & $30$M \\
\midrule
\multirow{2}{*}{\textit{detect-qwen-family}} 
  & Train & $429$K & $4.6$M & $100$M \\
  & Val   & $134$K & $1.4$M & $30$M \\
\midrule
\multirow{2}{*}{\textit{detect-falcon-family}} 
  & Train & $434$K & $4.5$M & $100$M \\
  & Val   & $180$K & $1.3$M & $30$M \\
\bottomrule
\end{tabular}
\caption{Dataset statistics for LLM family detection datasets}
\label{tab:llm_family_ds}
\end{table}

\subsection{Evaluation Pipeline}

We evaluated each experiment on the \textit{master-testset} dataset. For each evaluated model, standard binary classification metrics were reported. For experiments~\ref{ssec:exp1} and~\ref{ssec:exp2} we compute metrics per-token and for experiments~\ref{ssec:exp3} and~\ref{ssec:exp4} we compute metrics per-sample. By per-token evaluation, each token in the dataset is considered as an independent instance for classification. In per-sample metrics, we first aggregate the predictions for each token in a text sample into a single prediction. We then calculate the metrics based on these aggregated predictions and labels of the text samples.

\subsubsection{Ensemble-Based Evaluation Approach} \label{sssec:eval}

The evaluation of the ensemble-based approach was more complex because we needed to aggregate predictions from multiple detectors. First, for each selected sample from the evaluation dataset, we calculated the mean of predictions across all tokens for each detector. It resulted in 21 predictions per sample for the \textbf{Per LLM} approach and 6 predictions per sample for the \textbf{Per LLM Family} approach. For a selected sample, if all predictions are below 0.5, we classify that sample as human-written; otherwise, we classify it as AI-generated.

\subsubsection{External Model}

The \texttt{desklib/ai-text-detector-v1.01} model is a 430M parameter variant of \texttt{microsoft/deberta-v3-large}, fine-tuned specifically for AI-generated text detection~\cite{desklib_ai_text_detector_2025}. This model ranks fifth overall on the popular \href{https://raid-bench.xyz/leaderboard}{RAID benchmark}. Among the models available on Hugging Face, it is in first place. To compare the performance of our models against the established state-of-the-art open model, we evaluated this model on the \textit{master-testset} dataset. This model supports a context length of only $768$ tokens, which is over 10 times shorter than that of all our models. All texts exceeding this limit were truncated.

\section{Results} \label{sec:results}

We evaluated all experiments on the \textit{master-testset} dataset. We compared the performance of our models with the open source state-of-the-art model - \texttt{desklib/ai-text-detector-v1.01}.

\subsection{External Model}

We evaluated the \texttt{desklib/ai-text\allowbreak-detector-v1.01} on the \textit{master-testset} dataset. For this model, all metrics were computed per sample. Table~\ref{tab:external} presents the performance metrics. 

\begin{table}[H]
\centering
\small
\begin{tabular}{l|cccccc}
\toprule
\textbf{Eval Dataset} & \textbf{Acc.} & \textbf{Bal. Acc.} & \textbf{Prec.} & \textbf{Recall} & \textbf{F1} & \textbf{AUC} \\
\midrule
\textit{master-testset}         & 0.880 & 0.872 & 0.840 & 0.962 & 0.897 & 0.974\\
\bottomrule
\end{tabular}
\caption{Metrics of the external model on the \textit{master-testset} dataset}
\label{tab:external}
\end{table}

Overall, the model exhibits strong discriminative performance, with particularly high recall rates.

\subsection{LLM Fine-tuning for AI Text Detection}

We fine-tuned 16 selected local large language models (Table~\ref{tab:model_info}, excluding the four $70$B$+$ LLMs) on the \textit{master-large} training split. Table~\ref{tab:ft_eval} presents all calculated metrics for all fine-tuned models on the \textit{master-testset} dataset. For each metric, the best result is made bold, and the second-best result is underlined. All metrics were computed per token.
 
\begin{table}[H]
\centering
\small
\begin{tabular}{lcccccc}
\toprule
\textbf{Model} & \textbf{Loss} & \textbf{Acc.} & \textbf{B. Acc.} & \textbf{Prec.} & \textbf{Recall} & \textbf{F1} \\
\midrule
Llama-3.1-8B & \textbf{0.044} & 0.995 & 0.995 & 0.996 & 0.993 & 0.995 \\
Llama-3.2-3B & 0.051 & 0.994 & 0.994 & 0.995 & 0.992 & 0.994 \\
\midrule
Phi-3-mini & \underline{0.044} & 0.993 & 0.993 & 0.997 & 0.990 & 0.993 \\
Phi-3-small & 0.056 & \underline{0.995} & \underline{0.995} & 0.997 & \underline{0.993} & \underline{0.995} \\
Phi-3-medium & 0.048 & 0.995 & 0.995 & \underline{0.998} & 0.992 & 0.995 \\
Phi-3.5-mini & 0.046 & 0.993 & 0.993 & 0.996 & 0.990 & 0.993 \\
Phi-4-mini & 0.045 & 0.994 & 0.994 & 0.997 & 0.990 & 0.993 \\
Phi-4 & 0.046 & \textbf{0.996} & \textbf{0.996} & \textbf{0.998} & \textbf{0.993} & \textbf{0.996} \\
\midrule
Ministral-8B-2410 & 0.048 & 0.994 & 0.994 & 0.996 & 0.992 & 0.994 \\
Mistral-Nemo-2407 & 0.054 & 0.994 & 0.994 & 0.997 & 0.990 & 0.993 \\
\midrule
Qwen2-7B & 0.048 & 0.994 & 0.994 & 0.997 & 0.991 & 0.994 \\
Qwen2.5-3B & 0.052 & 0.993 & 0.993 & 0.996 & 0.991 & 0.993 \\
Qwen2.5-7B & 0.051 & 0.993 & 0.993 & 0.996 & 0.991 & 0.993 \\
Qwen2.5-14B & 0.051 & 0.993 & 0.993 & 0.996 & 0.990 & 0.993 \\
\midrule
Falcon3-3B & 0.051 & 0.993 & 0.993 & 0.995 & 0.991 & 0.993 \\
Falcon3-7B & 0.050 & 0.994 & 0.994 & 0.996 & 0.991 & 0.994 \\
\bottomrule
\end{tabular}
\caption{Performance metrics of all fine-tuned models on the \textit{master-testset} dataset}
\label{tab:ft_eval}
\end{table}

All models perform substantially better than the external model. The best performing fine-tuned LLM is the \texttt{Phi-4}, which achieves the highest scores across all but one metric. The second-best model is \texttt{Phi-3-small}, which places second on four of six metrics. Overall, the differences between model performances are minimal.

Notably, all models achieve near-perfect token-level accuracy, indicating robust discriminative capability. However, this may partly reflect the controlled nature of the generated datasets, and real-world performance could be lower when facing novel or heavily paraphrased AI-generated text.

\subsection{LLM Fine-tuning for Proprietary Model Detection}

We fine-tuned 16 large language models (Table~\ref{tab:model_info}, excluding the four $70$B$+$ LLMs) to detect texts generated by the \texttt{gpt-4.1-nano-2025-04-14} model. Table~\ref{tab:gpt_eval} presents metrics for all fine-tuned models on the validation split of the \textit{detect-gpt-4.1-nano} dataset. For each LLM, we selected the highest validation accuracy value achieved during training. These models were not evaluated on the \textit{master-testset}, as the best-performing detector is included in the ensembles evaluated in the subsequent experiments.

\begin{table}[H]
\centering
\small
\begin{tabular}{lcccccc}
\toprule
\textbf{Model} & \textbf{Loss} & \textbf{Acc.} & \textbf{B. Acc.} & \textbf{Prec.} & \textbf{Recall} & \textbf{F1} \\
\midrule
Llama-3.1-8B & 0.229 & 0.937 & 0.937 & 0.930 & 0.946 & 0.938 \\
Llama-3.2-3B & 0.264 & 0.931 & 0.931 & 0.932 & 0.930 & 0.931 \\
\midrule
Phi-3-mini & 0.248 & 0.929 & 0.929 & 0.922 & 0.939 & 0.930 \\
Phi-3-small & \underline{0.217} & 0.939 & 0.939 & 0.927 & \underline{0.953} & 0.940 \\
Phi-3-medium & 0.218 & 0.938 & 0.938 & 0.928 & 0.950 & 0.939 \\
Phi-3.5-mini & 0.268 & 0.924 & 0.924 & 0.923 & 0.926 & 0.925 \\
Phi-4-mini & 0.251 & 0.923 & 0.923 & 0.909 & 0.939 & 0.924 \\
Phi-4 & \textbf{0.199} & \textbf{0.947} & \textbf{0.947} & \textbf{0.939} & \textbf{0.956} & \textbf{0.947} \\
\midrule
Ministral-8B-2410 & 0.227 & 0.936 & 0.936 & 0.928 & 0.946 & 0.937 \\
Mistral-Nemo-2407 & 0.222 & \underline{0.940} & \underline{0.940} & \underline{0.936} & 0.945 & \underline{0.940} \\
\midrule
Qwen2-7B & 0.257 & 0.929 & 0.929 & 0.929 & 0.929 & 0.929 \\
Qwen2.5-3B & 0.280 & 0.922 & 0.922 & 0.920 & 0.924 & 0.922 \\
Qwen2.5-7B & 0.246 & 0.929 & 0.929 & 0.918 & 0.940 & 0.929 \\
Qwen2.5-14B & 0.227 & 0.935 & 0.935 & 0.930 & 0.939 & 0.935 \\
\midrule
Falcon3-3B & 0.286 & 0.918 & 0.918 & 0.917 & 0.917 & 0.917 \\
Falcon3-7B & 0.291 & 0.913 & 0.913 & 0.910 & 0.913 & 0.912 \\
\bottomrule
\end{tabular}
\caption{Performance metrics of all models fine-tuned for proprietary model detection on the \textit{detect-gpt-4.1-nano} dataset validation split}
\label{tab:gpt_eval}
\end{table}

The \texttt{Phi-4} model achieves the best overall performance, obtaining the highest scores across all evaluation metrics, with an accuracy of $0.947$. The \texttt{Mistral-Nemo-2407} model ranks second, placing second on four of the six metrics. Based on these results, the fine-tuned \texttt{Phi-4} model was selected and included in both the \textbf{Per LLM} and \textbf{Per LLM Family} ensembles.

\subsection{Per LLM Fine-Tuning for AI Text Detection}

Table~\ref{tab:perllm_eval} presents metrics for all models fine-tuned for the detection of texts generated by themselves on respective datasets' validation splits. For each LLM, we selected the highest validation accuracy during training.

\begin{table}[H]
\centering
\small
\begin{tabular}{lrrrrrr}
\toprule
\textbf{Model} & \textbf{Loss} & \textbf{Acc.} & \textbf{B. Acc.} & \textbf{Prec.} & \textbf{Recall} & \textbf{F1} \\
\midrule
Llama-3.1-8B & 0.417 & 0.852 & 0.852 & 0.821 & 0.904 & 0.861 \\
Llama-3.1-70B-AWQ & 0.331 & 0.873 & 0.874 & 0.845 & 0.908 & 0.876 \\
Llama-3.2-3B & 0.387 & 0.877 & 0.877 & 0.872 & 0.891 & 0.881 \\
Llama-3.3-70B-AWQ & 0.333 & 0.875 & 0.873 & 0.847 & 0.910 & 0.876 \\
\midrule
Phi-3-mini & 0.478 & 0.807 & 0.806 & 0.788 & 0.843 & 0.815 \\
Phi-3-small & 0.365 & 0.851 & 0.851 & 0.852 & 0.857 & 0.855 \\
Phi-3-medium & 0.328 & 0.858 & 0.858 & 0.850 & 0.861 & 0.855 \\
Phi-3.5-mini & \underline{0.277} & \underline{0.908} & \underline{0.908} & \underline{0.906} & \underline{0.916} & \underline{0.911} \\
Phi-4-mini & 0.520 & 0.778 & 0.777 & 0.745 & 0.849 & 0.794 \\
phi-4 & 0.366 & 0.888 & 0.888 & 0.877 & 0.898 & 0.888 \\
\midrule
Ministral-8B-2410 & 0.343 & 0.877 & 0.877 & 0.865 & 0.891 & 0.878 \\
Mistral-Nemo-2407 & \textbf{0.273} & \textbf{0.923} & \textbf{0.923} & \textbf{0.920} & \textbf{0.925} & \textbf{0.922} \\
\midrule
Qwen2-7B & 0.431 & 0.848 & 0.848 & 0.855 & 0.848 & 0.851 \\
Qwen2-72B-AWQ & 0.321 & 0.863 & 0.864 & 0.835 & 0.898 & 0.866 \\
Qwen2.5-3B & 0.455 & 0.827 & 0.827 & 0.813 & 0.852 & 0.832 \\
Qwen2.5-7B & 0.489 & 0.803 & 0.802 & 0.794 & 0.834 & 0.813 \\
Qwen2.5-14B & 0.439 & 0.845 & 0.844 & 0.834 & 0.864 & 0.849 \\
Qwen2.5-72B-AWQ & 0.323 & 0.865 & 0.866 & 0.834 & 0.896 & 0.867 \\
\midrule
Falcon3-3B & 0.496 & 0.821 & 0.822 & 0.809 & 0.832 & 0.820 \\
Falcon3-7B & 0.501 & 0.816 & 0.816 & 0.813 & 0.828 & 0.820 \\
\bottomrule
\end{tabular}
\caption{Performance metrics of all models fine-tuned for the detection of texts generated by themselves on the respective datasets' validation splits}
\label{tab:perllm_eval}
\end{table}

The \texttt{Mistral-Nemo-2407} model achieves the best overall performance in self-detection, obtaining the highest values across all evaluation metrics. The \texttt{Phi-3.5-mini} model ranks second, earning the second-highest scores across all metrics.

We evaluate the ensemble as described in the section~\ref{sssec:eval}. Each sample is associated with 21 model predictions. For a selected sample, if all predictions are below 0.5, we classify that sample as human-written; otherwise, we classify it as AI-generated. All metrics were computed per sample. Table~\ref{tab:ensamble_hard_perllm} presents metrics of the \textbf{Per LLM} ensemble on the \textit{master-testset} dataset.

\begin{table}[H]
\centering
\small
\begin{tabular}{l|cccccc}
\toprule
\textbf{Eval Dataset} & \textbf{Acc.} & \textbf{B. Acc.} & \textbf{Prec.} & \textbf{Recall} & \textbf{F1} & \textbf{AUC} \\
\midrule
\textit{master-testset}         & 0.713 & 0.687 & 0.656 & 0.988 & 0.788 & 0.791 \\
\bottomrule
\end{tabular}
\caption{Performance metrics of the \textbf{Per LLM} ensemble on the \textit{master-testset} dataset}
\label{tab:ensamble_hard_perllm}
\end{table}

Overall, the \textbf{Per LLM} ensemble demonstrates moderate performance. It underperforms in comparison with the external model and individual LLMs fine-tuned for AI text detection, suggesting that the extreme computational cost of running 21 detectors is not justified by any performance gains.

The ensemble achieves a high recall of $0.98$ but a lower precision of $0.66$. This is likely because aggregating predictions from 21 detectors increases the likelihood that at least one detector will misclassify human-written text as AI-generated, thereby inflating recall while reducing precision.

\subsection{Per LLM Family Fine-Tuning for AI Text Detection}

\begin{figure}[H]
    \centering
    \includegraphics[width=0.9\textwidth]{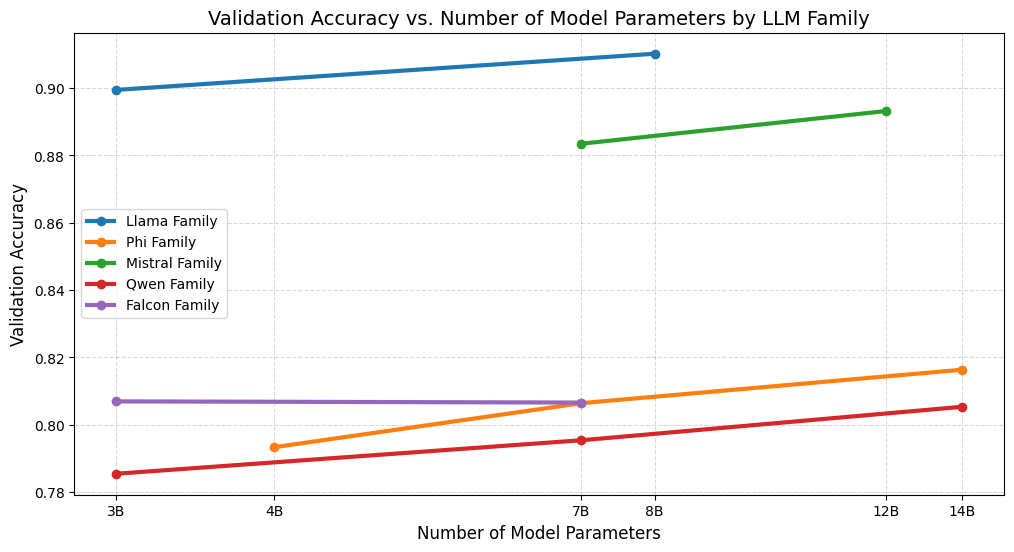}
    \caption{Validation accuracy vs number of model parameters by LLM family}
    \label{fig:perllmfamily_nice}
\end{figure}

We fine-tuned 16 selected local large language models (Table~\ref{tab:model_info}, excluding the four $70$B$+$ LLMs) to detect texts from all models within their respective families. Figure~\ref{fig:perllmfamily_nice} presents validation accuracy vs. the number of model parameters by LLM family.

Across all families, performance increases only modestly with model size. Within each family, the largest model consistently achieves the highest validation accuracy. Among the families, \texttt{Llama} performs best, while \texttt{Qwen} performs worst. Notably, the two top-performing families, \texttt{Llama} and \texttt{Mistral}, clearly outperform the remaining three families.

Table~\ref{tab:perllmfamily_eval} presents metrics of the best model in each family fine-tuned for the detection of texts generated by all LLMs within their families on the respective datasets’ validation splits.

\begin{table}[H]
\centering
\small
\begin{tabular}{lcccccc}
\toprule
\textbf{Model} & \textbf{Loss} & \textbf{Acc.} & \textbf{B. Acc.} & \textbf{Prec.} & \textbf{Recall} & \textbf{F1} \\
\midrule
Llama-3.1-8B & \textbf{0.318} & \textbf{0.910} & \textbf{0.910} & \textbf{0.910} & \textbf{0.913} & \textbf{0.911} \\
Phi-4 & 0.457 & 0.825 & 0.825 & 0.837 & 0.800 & 0.818 \\
Mistral-Nemo-2407 & \underline{0.320} & \underline{0.893} & \underline{0.893} & \underline{0.903} & \underline{0.877} & \underline{0.890} \\
Qwen2.5-14B & 0.509 & 0.805 & 0.805 & 0.811 & 0.800 & 0.805 \\
Falcon3-7B & 0.513 & 0.807 & 0.807 & 0.799 & 0.818 & 0.808 \\
\bottomrule
\end{tabular}
\caption{Performance metrics of the best model in each family fine-tuned for the detection of texts generated by LLMs from their families on the respective datasets' validation splits}
\label{tab:perllmfamily_eval}
\end{table}

Within each LLM family, the largest model achieves the highest performance. Overall, \texttt{Llama-3.1-8B} ranks first across all metrics, followed by \texttt{Mistral-Nemo-2407} in second place—models from the \texttt{Qwen} and \texttt{Falcon} families show lower performance.

We evaluate the ensemble as described in the section~\ref{sssec:eval}. Each sample is associated with 6 model predictions. For a selected sample, if all predictions are below 0.5, we classify that sample as human-written; otherwise, we classify it as AI-generated. All metrics were computed per sample. Table~\ref{tab:ensamble_hard_perllmfamily} presents metrics of the \textbf{Per LLM Family} ensemble on the \textit{master-testset} dataset.

\begin{table}[H]
\centering
\small
\begin{tabular}{l|cccccc}
\toprule
\textbf{Eval Dataset} & \textbf{Acc.} & \textbf{B. Acc.} & \textbf{Prec.} & \textbf{Recall} & \textbf{F1 Score} & \textbf{AUC} \\
\midrule
\textit{master-testset}         & 0.806 & 0.792 & 0.751 & 0.961 & 0.843 & 0.867 \\
\bottomrule
\end{tabular}
\caption{Performance metrics of the \textbf{Per LLM Family} ensemble}
\label{tab:ensamble_hard_perllmfamily}
\end{table}

Overall, the \textbf{Per LLM Family} ensemble performs better than the \textbf{Per LLM} ensemble. It achieves a much higher score on all reported metrics. Moreover, it utilizes only 6 detectors instead of 21, thereby significantly reducing inference costs.

\subsection{Summary of Results}

\begin{figure}[H]
    \begin{center}
        \includegraphics[width=0.95\textwidth]{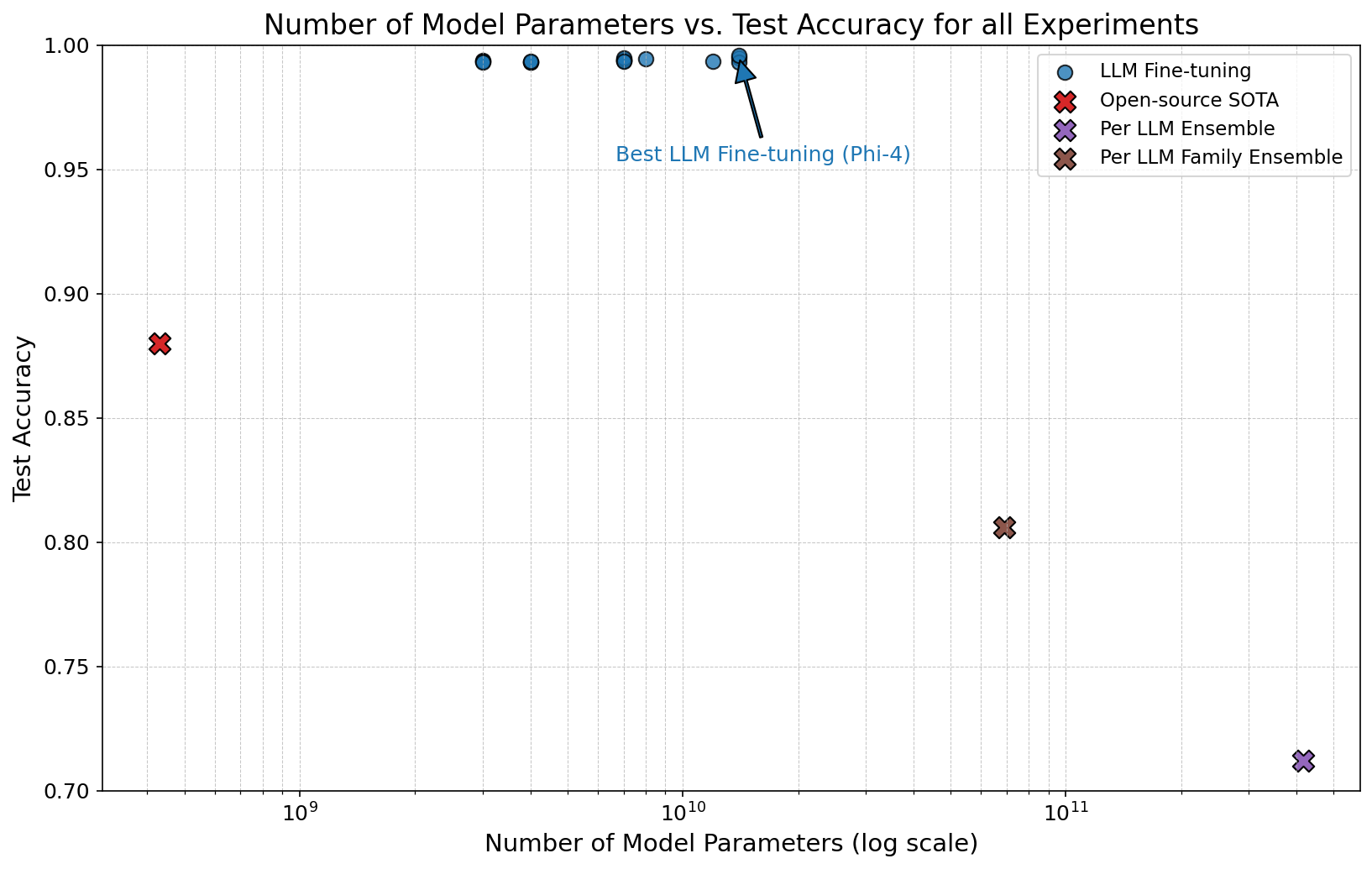}
        \caption{Scatter plot of number of model parameters vs accuracy on the \textit{master-testset} dataset for all conducted experiments}  
        \label{fig:all_results}
    \end{center}
\end{figure}

Figure~\ref{fig:all_results} visualizes the relationship between model size and accuracy across all experiments, highlighting the trade-off between parameter count and performance. For ensembles, the parameter count represents the sum of parameters of all LLMs in the ensemble.

The fine-tuned \texttt{Phi-4} model achieves the highest performance, with near-perfect accuracy. Both ensembles underperform relative to the open-source SOTA model, with the \textbf{Per LLM} ensemble achieving the lowest accuracy despite having the largest number of parameters.

Table~\ref{tab:results_last} summarizes the performance metrics and parameter counts for the best-performing models from each experiment, as well as the open-source \texttt{desklib/ai-text-detector-v1.01} model on the \textit{master-testset} dataset. 

\begin{table}[H]
\centering
\footnotesize
\resizebox{\textwidth}{!}{%
\begin{tabular}{lc|cccccc}
\toprule
\textbf{Model} & \textbf{\# Params} & \textbf{Acc.} & \textbf{B. Acc.} & \textbf{Prec.} & \textbf{Recall} & \textbf{F1} \\
\midrule
\texttt{Open-source SOTA}~\cite{desklib_ai_text_detector_2025} & 430M & 0.880 & 0.872 & 0.840 & 0.962 & 0.897 \\
\texttt{Best LLM Finetune} & 14.7B & \textbf{0.996} & \textbf{0.996} & \textbf{0.998} & \textbf{0.993} & \textbf{0.996} \\
\texttt{Per LLM Ensemble} & 416B & 0.713 & 0.687 & 0.656 & 0.988 & 0.788 \\
\texttt{Per LLM Family Ensemble} & 69B & 0.806 & 0.792 & 0.751 & 0.961 & 0.843 \\
\bottomrule
\end{tabular}
}
\caption{Performance metrics of best-performing models from each conducted experiment on the \textit{master-testset} dataset}
\label{tab:results_last}
\end{table}

The fine-tuned \texttt{Phi-4} model achieves the highest scores across all metrics, approaching perfect accuracy. Both ensembles achieve significantly higher recall than precision. This pattern reflects their reliance on multiple detectors: aggregating predictions increases the likelihood of classifying human-written text as AI-generated, boosting recall while reducing precision. Compared to the open-source SOTA model, \texttt{Phi-4} substantially outperforms in all metrics, while both ensembles lag in overall accuracy despite their larger parameter counts.

\section{Conclusion} \label{sec:conclusion}

A comprehensive series of experiments was conducted to evaluate the performance of various AI-generated text detection methods. The initial phase of development involved constructing \textit{master} datasets to support consistent training and evaluation across multiple detection methods. Traditional transformer-based models were implemented to establish strong performance baselines. The article further introduced two novel fine-tuning paradigms - \textbf{Per LLM} and \textbf{Per LLM family}, designed to improve detection accuracy and generalizability. 

Several of our fine-tuned models outperform the open-source state-of-the-art detector and support context lengths over ten times larger. The best-performing model achieves near-perfect token-level accuracy on a 100-million-token benchmark, demonstrating the potential of fine-tuned LLMs to reliably distinguish AI-generated from human-written text under controlled conditions. The \textbf{Per LLM} and \textbf{Per LLM Family} ensembles, while not surpassing the overall accuracy of the best baseline, offer the unique ability to pinpoint the specific generative model (family) responsible for each text sample, providing fine-grained detection insights. Their performance may improve if AI-generated text were produced in ways that more closely reflect real-world conditions, such as hybrid human-AI content or diverse prompting strategies. Overall, these results highlight the effectiveness of our proposed framework and training paradigms, while emphasizing the need for further evaluation under more diverse and realistic conditions.

\section{Limitations and Future Work} \label{sec:limitations}

Despite the strong performance of our models, several limitations remain. First, the AI-generated texts were produced by prompting human-written samples, and uniform sampling parameters were applied across all LLMs. This may limit the diversity of generated texts and the generalizability of detection models to more realistic or adversarial scenarios, such as hybrid human-AI content or varied prompting strategies. Second, our experiments primarily used relatively small LLMs and datasets compared with current large-scale models; scaling both model size and training data may further improve performance and robustness. Third, while the \textbf{Per LLM} and \textbf{Per LLM Family} paradigms enable the identification of a text's specific generative model, their applicability to unseen models or real-world AI-generated content remains to be evaluated. Finally, the methodologies presented here could be extended beyond text to multi-modal AI content, including images, video, and audio, offering opportunities for developing unified, multi-modal detection frameworks. Addressing these limitations in future work will be crucial for developing detection systems that are both highly accurate and robust in real-world conditions.

\newpage

\bibliographystyle{elsarticle-num} 
\bibliography{references.bib}




% %% For numbered reference style

%   \textit{\LaTeX: a document preparation system},
%   Addison Wesley, Massachusetts,

% \end{thebibliography}

\appendix
\section{Data Sources} \label{app:ds}

We provide the links to human datasets used in our research:
\begin{itemize}
    \item Tweets - \url{https://www.kaggle.com/datasets/kazanova/sentiment140}
    \item Reddit - \url{https://www.kaggle.com/datasets/smagnan/1-million-reddit-comments-from-40-subreddits}
    \item Blogs - \url{https://www.kaggle.com/datasets/rtatman/blog-authorship-corpus}
    \item NYT Articles and Comments - \url{https://www.kaggle.com/datasets/benjaminawd/new-york-times-articles-comments-2020}
    \item Essays - \url{https://www.kaggle.com/datasets/thedrcat/daigt-external-train-dataset}
    \item RAID - \url{https://huggingface.co/datasets/liamdugan/raid}
    \item XSum - \url{https://huggingface.co/datasets/EdinburghNLP/xsum}
    \item WritingPrompts - \url{https://huggingface.co/datasets/euclaise/writingprompts}
    \item Natural Questions - \url{https://huggingface.co/datasets/google-research-datasets/natural_questions}
\end{itemize}

\section{Prompt Templates} \label{app:pt}

We provide prompt templates for all datasets in Tables~\ref{tab:base_prompt_blogs} through~\ref{tab:base_prompt_xsum}.

\begin{table}[H]
    \centering
    \begin{tabular}{|l|p{10.5cm}|}
        \hline
        \multicolumn{2}{|c|}{\textbf{Dataset: Blogs}} \\
        \hline
        \textbf{Role} & \textbf{Content} \\
        \hline
        System & You are a helpful assistant for rewriting blogs. Based on the provided blog, generate a similar one. MAKE SURE TO REPLY ONLY WITH THE SIMILAR BLOG. \\
        \hline
        User & Blog:\textbackslash n\{blog\} \\
        \hline
        Assistant & Similar blog:\textbackslash n \\
        \hline
    \end{tabular}
    \caption{Prompt template for the Blogs dataset}
    \label{tab:base_prompt_blogs}
\end{table}

\begin{table}[H]
    \centering
    \begin{tabular}{|l|p{10.5cm}|}
        \hline
        \multicolumn{2}{|c|}{\textbf{Dataset: Essays}} \\
        \hline
        \textbf{Role} & \textbf{Content} \\
        \hline
        System & You are a helpful assistant for rewriting students' essays. Based on the provided essay, generate a similar one in a natural and authentic tone, maintaining the same meaning but rephrased. Ensure the rewritten essay matches the length of the original, and avoids overly formal or advanced phrasing. MAKE SURE TO REPLY ONLY WITH THE SIMILAR ESSAY. \\
        \hline
        User & Essay:\textbackslash n\{essay\} \\
        \hline
        Assistant & Similar essay:\textbackslash n \\
        \hline
    \end{tabular}
    \caption{Prompt template for the Essays dataset}
    \label{tab:base_prompt_essays}
\end{table}

\begin{table}[H]
    \centering
    \begin{tabular}{|l|p{10.5cm}|}
        \hline
        \multicolumn{2}{|c|}{\textbf{Dataset: Natural Questions}} \\
        \hline
        \textbf{Role} & \textbf{Content} \\
        \hline
        System & You are a helpful assistant for answering questions based on the provided context. The context will be a copy of a Wikipedia article. Answer the question based only on the given context. MAKE SURE TO REPLY ONLY WITH THE ANSWER. \\
        \hline
        User & Context:\textbackslash n\{context\}\textbackslash nQuestion: \{question\} \\
        \hline
        Assistant & Answer:\textbackslash n \\
        \hline
    \end{tabular}
    \caption{Prompt template for the Natural Questions dataset}
    \label{tab:base_prompt_natural_questions}
\end{table}

\begin{table}[H]
    \centering
    \begin{tabular}{|l|p{10.5cm}|}
        \hline
        \multicolumn{2}{|c|}{\textbf{Dataset: NYT Articles}} \\
        \hline
        \textbf{Role} & \textbf{Content} \\
        \hline
        System & You are a helpful assistant for writing article abstracts. Based on the provided headline and list of keywords, generate an abstract for the article. Ensure the abstract maintains a similar length to typical article abstracts. MAKE SURE TO REPLY ONLY WITH THE ABSTRACT. \\
        \hline
        User & Headline:\textbackslash n\{headline\}\textbackslash nKeywords:\textbackslash n\{keywords\} \\
        \hline
        Assistant & Abstract:\textbackslash n \\
        \hline
    \end{tabular}
    \caption{Prompt template for the NYT Articles dataset}
    \label{tab:base_prompt_nyt_articles}
\end{table}

\begin{table}[H]
    \centering
    \begin{tabular}{|l|p{10.5cm}|}
        \hline
        \multicolumn{2}{|c|}{\textbf{Dataset: NYT Comments}} \\
        \hline
        \textbf{Role} & \textbf{Content} \\
        \hline
        System & You are a helpful assistant for writing comments based on article abstracts and sample comments. Based on the provided article abstract and sample comment, generate a similar comment related to the article. Ensure the comment matches the tone and length of the sample comment. MAKE SURE TO REPLY ONLY WITH THE COMMENT. \\
        \hline
        User & Abstract:\textbackslash n\{abstract\}\textbackslash nComment:\textbackslash n\{comment\} \\
        \hline
        Assistant & Similar comment:\textbackslash n \\
        \hline
    \end{tabular}
    \caption{Prompt template for the NYT Comments dataset}
    \label{tab:base_prompt_nyt_comments}
\end{table}

\begin{table}[H]
    \centering
    \begin{tabular}{|l|p{10.5cm}|}
        \hline
        \multicolumn{2}{|c|}{\textbf{Dataset: RAID}} \\
        \hline
        \textbf{Role} & \textbf{Content} \\
        \hline
        System & You are a helpful assistant specializing in writing texts across various domains, including abstracts and news articles, based on provided titles. Based on the given domain and title, generate a text of appropriate length and style that aligns with the specified domain. MAKE SURE TO REPLY ONLY WITH THE GENERATED TEXT. \\
        \hline
        User & Domain:\textbackslash n\{domain\}\textbackslash nTitle:\textbackslash n\{title\} \\
        \hline
        Assistant & Generated text:\textbackslash n \\
        \hline
    \end{tabular}
    \caption{Prompt template for the RAID dataset}
    \label{tab:base_prompt_raid}
\end{table}

\begin{table}[H]
    \centering
    \begin{tabular}{|l|p{10.5cm}|}
        \hline
        \multicolumn{2}{|c|}{\textbf{Dataset: Reddit}} \\
        \hline
        \textbf{Role} & \textbf{Content} \\
        \hline
        System & You are a helpful assistant for rewriting Reddit comments. Based on the provided comment and the subreddit where it was posted, generate a similar comment that fits the context of the subreddit. MAKE SURE TO REPLY ONLY WITH THE SIMILAR COMMENT. \\
        \hline
        User & Comment:\textbackslash n\{comment\}\textbackslash nSubreddit:\textbackslash n\{subreddit\} \\
        \hline
        Assistant & Similar comment:\textbackslash n \\
        \hline
    \end{tabular}
    \caption{Prompt template for the Reddit dataset}
    \label{tab:base_prompt_reddit}
\end{table}

\begin{table}[H]
    \centering
    \begin{tabular}{|l|p{10.5cm}|}
        \hline
        \multicolumn{2}{|c|}{\textbf{Dataset: Tweets}} \\
        \hline
        \textbf{Role} & \textbf{Content} \\
        \hline
        System & You are a helpful assistant for rewriting tweets. Based on the provided tweet, generate a similar one while maintaining the original meaning and tone. MAKE SURE TO REPLY ONLY WITH THE SIMILAR TWEET. \\
        \hline
        User & Tweet:\textbackslash n\{tweet\} \\
        \hline
        Assistant & Similar tweet:\textbackslash n \\
        \hline
    \end{tabular}
    \caption{Prompt template for the Tweets dataset}
    \label{tab:base_prompt_tweets}
\end{table}

\begin{table}[H]
    \centering
    \begin{tabular}{|l|p{10.5cm}|}
        \hline
        \multicolumn{2}{|c|}{\textbf{Dataset: WritingPrompts}} \\
        \hline
        \textbf{Role} & \textbf{Content} \\
        \hline
        System & You are a helpful assistant for writing stories based on the provided prompt. Based on the given prompt, generate a story that aligns with it. MAKE SURE TO REPLY ONLY WITH THE STORY. \\
        \hline
        User & Prompt:\textbackslash n\{prompt\} \\
        \hline
        Assistant & Story:\textbackslash n \\
        \hline
    \end{tabular}
    \caption{Prompt template for the WritingPrompts dataset}
    \label{tab:base_prompt_writingprompts}
\end{table}

\begin{table}[H]
    \centering
    \begin{tabular}{|l|p{10.5cm}|}
        \hline
        \multicolumn{2}{|c|}{\textbf{Dataset: XSum}} \\
        \hline
        \textbf{Role} & \textbf{Content} \\
        \hline
        System & You are a helpful assistant for writing news articles based on the provided one-sentence summaries. Based on the given summary, generate a full news article. MAKE SURE TO REPLY ONLY WITH THE NEWS ARTICLE. \\
        \hline
        User & Summary:\textbackslash n\{summary\} \\
        \hline
        Assistant & News article:\textbackslash n \\
        \hline
    \end{tabular}
    \caption{Prompt template for the XSum dataset}
    \label{tab:base_prompt_xsum}
\end{table}

\section{Training Setup Details} \label{app:tsd}

Table~\ref{tab:ft1_hp} presents the selected hyperparameters for fine-tuning LLMs in this and all of the following experiments.

\begin{table}[H]
\centering
\begin{tabular}{@{}lccc@{}}
\toprule
\textbf{Model} & \textbf{Min LR} & \textbf{Max LR} & \textbf{Batch Size} \\
\midrule
Llama-3.1-8B-Instruct & 2e-5 & 2e-4 & 64 \\
Meta-Llama-3.1-70B-Instruct & 1e-5 & 1e-4 & 16 \\
Llama-3.2-3B-Instruct & 5e-5 & 5e-4 & 128 \\
Meta-Llama-3.3-70B-Instruct & 1e-5 & 1e-4 & 16 \\
\midrule
Phi-3-mini-128k-instruct & 3e-5 & 3e-4 & 128 \\
Phi-3-small-128k-instruct & 2e-5 & 2e-4 & 64 \\
Phi-3-medium-128k-instruct & 1e-5 & 1e-4 & 32 \\
Phi-3.5-mini-instruct & 3e-5 & 3e-4 & 128 \\
Phi-4-mini-instruct & 3e-5 & 3e-4 & 128 \\
Phi-4 & 1e-5 & 1e-4 & 32 \\
\midrule
Ministral-8B-Instruct-2410 & 2e-5 & 2e-4 & 64 \\
Mistral-Nemo-Instruct-2407 & 2e-5 & 2e-4 & 64 \\
\midrule
Qwen2-7B-Instruct & 2e-5 & 2e-4 & 64 \\
Qwen2-72B-Instruct-AWQ & 1e-5 & 1e-4 & 16 \\
Qwen2.5-3B-Instruct & 5e-5 & 5e-4 & 128 \\
Qwen2.5-7B-Instruct & 2e-5 & 2e-4 & 64 \\
Qwen2.5-14B-Instruct & 1e-5 & 1e-4 & 32 \\
Qwen2.5-72B-Instruct-AWQ & 1e-5 & 1e-4 & 16 \\
\midrule
Falcon3-3B-Instruct & 5e-5 & 5e-4 & 128 \\
Falcon3-7B-Instruct & 2e-5 & 2e-4 & 64 \\
\bottomrule
\end{tabular}
\caption{Hyperparameters for fine-tuning LLMs}
\label{tab:ft1_hp}
\end{table}

\begin{figure}[H]
    \centering
    \includegraphics[width=0.8\textwidth]{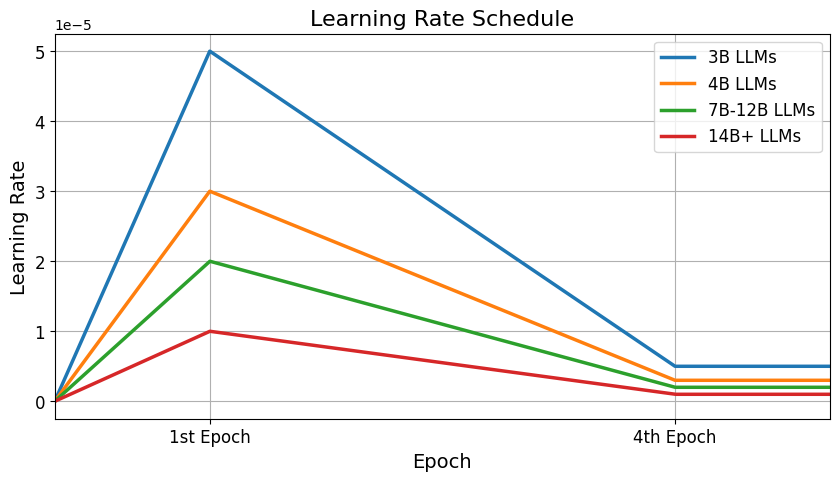}
    \caption{Learning rate schedule for fine-tuning LLMs}
    \label{fig:ft_lr_sched}
\end{figure}

We used a linear learning rate schedule with a $20\%$ warm-up (1 epoch). The learning rate starts at the \texttt{Min LR} value and linearly increases over one epoch to the \texttt{Max LR} value. After that, it follows a linear decay, so that in the fourth epoch, it again reaches the \texttt{Min LR} value. Then it stays constant for the last epoch. Figure~\ref{fig:ft_lr_sched} presents the learning rate value through the fine-tuning process.

\end{document}